# How Did Humans Become So Creative? A Computational Approach

**Liane Gabora**
Department of Psychology
University of British Columbia, 3333 University Way
Kelowna BC, CANADA, V1V 1V7
liane.gabora@ubc.ca

**Steve DiPaola**
Department of Cognitive Science / SIAT
Simon Fraser University, 250-13450 102 Ave
Surrey BC, CANADA, V3T 0A3
sdipaola@sfu.ca

**Abstract**

This paper summarizes efforts to computationally model two transitions in the evolution of human creativity: its origins about two million years ago, and the 'big bang' of creativity about 50,000 years ago. Using a computational model of cultural evolution in which neural network based agents evolve ideas for actions through invention and imitation, we tested the hypothesis that human creativity began with onset of the capacity for recursive recall. We compared runs in which agents were limited to single-step actions to runs in which they used recursive recall to chain simple actions into complex ones. Chaining resulted in higher diversity, open-ended novelty, no ceiling on the mean fitness of actions, and greater ability to make use of learning. Using a computational model of portrait painting, we tested the hypothesis that the explosion of creativity in the Middle/Upper Paleolithic was due to onset of contextual focus: the capacity to shift between associative and analytic thought. This resulted in faster convergence on portraits that resembled the sitter, employed painterly techniques, and were rated as preferable. We conclude that recursive recall and contextual focus provide a computationally plausible explanation of how humans evolved the means to transform this planet.

## Introduction

To gain insight into the mechanisms underlying creativity, one might start by testing peoples' creative abilities, perhaps using technologies such as fMRI, or dissect the brains of people who were known to be particularly creative during their lifetimes. However, to gain insight into the *evolution* of creativity, these options do not exist. All that is left of our prehistoric ancestors are their bones and artifacts such as stone tools that resist the passage of time. Thus to understand the evolution of creativity, computational modeling is virtually the only scientific tool we have.

Humans are not only creative; we put our own spin on the inventions of others, such that new inventions build cumulatively on previous ones. This cumulative cultural change is referred to as the *ratchet effect* (Tomasello, Kruger, & Ratner, 1993), and it has been suggested that it is uniquely human (Donald, 1991). A mathematical model of two transitions in the evolution of the cognitive mechanisms underlying creativity has been put forward (Gabora & Aerts, 2009). Computational models of these mechanisms have also been developed (DiPaola & Gabora, 2007, 2009; Gabora, 1995, 2008a,b; Gabora & Leijnen, 2009; Leijnen & Gabora, 2009, 2010; Gabora & Saberi, 2011). However, these efforts used different modeling platforms, and because the aims underlying them have been part scientific and part artistic, their relevance to each other, and to an overarching research program has not previously been made clear. The goal of this paper is to explain how, together, they constitute an integrated effort to computationally model the evolution of human creativity.

## First Transition: The Earliest Signs of Creativity

The minds of our earliest ancestors, *Homo habilis,* have been referred to as *episodic* because there is no evidence that they deviated from the present moment of concrete sensations (Donald, 1991). They could encode perceptions of events in memory, and recall them in the presence of a cue, but had little voluntary access to memories without environmental cues. They were therefore unable to shape, modify, or practice skills and actions, and unable to invent or refine complex gestures or vocalizations.

*Homo erectus* lived between approximately 1.8 and 0.3 million years ago. The cranial capacity of the *Homo erectus* brain was approximately 1,000 cc, about 25% larger than that of *Homo habilis*, at least twice as large as that of living great apes, and 75% that of modern humans (Ruff et al., 1997). This period is widely referred to as the beginnings of cumulative culture. *Homo erectus* exhibited many indications of enhanced intelligence, creativity, and ability to adapt to their environment, including sophisticated, task-specific stone hand axes, complex stable seasonal home bases, and long-distance hunting strategies involving large game, and migration out of Africa.

This period marks the onset of the archaeological record and it is thought to be the beginnings of human culture. It is widely believed that this cultural transition reflects an underlying transition in cognitive or social abilities. Some have suggested that they owe their achievements to onset of *theory of mind* (Mithen, 1998) or the capacity to imitate (Dugatkin, 2001). However, there is evidence that other species possess theory of mind and the capacity to imitate (Heyes, 1998), yet do not compare to modern humans in intelligence and cultural complexity.

Evolutionary psychologists have suggested that the intelligence and cultural complexity of the *Homo* line is due to the onset of *massive modularity* (Buss, 1999, 2004; Barkow, Cosmides, &Tooby, 1992). However, although the mind exhibits an intermediate degree of functional and anatomical modularity, neuroscience has not revealed vast numbers of hardwired, encapsulated, task-specific modules; indeed, the brain has been shown to be more highly subject to environmental influence than was previously believed (Buller, 2005; Byrne, 2000; Wexler, 2006).

## A Promising and Testable Hypothesis

Donald (1991) proposed that with the enlarged cranial capacity of *Homo erectus*, the human mind underwent the first of three transitions by which it—along with the cultural matrix in which it is embedded—evolved from the ancestral, pre-human condition. This transition is characterized by a shift from an *episodic* to a *mimetic mode* of cognitive functioning, made possible by onset of the capacity for voluntary retrieval of stored memories, independent of environmental cues. Donald refers to this as a *self-triggered recall and rehearsal loop*. Self-triggered recall enabled information to be processed recursively with respect to different contexts or perspectives. It allowed our ancestor to access memories voluntarily and thereby act out[1] events that occurred in the past or that might occur in the future. Thus not only could the mimetic mind temporarily escape the here and now, but by miming or gesture, it could communicate similar escapes in other minds. The capacity to mime thus ushered forth what is referred to as a *mimetic* form of cognition and brought about a transition to the mimetic stage of human culture. The self-triggered recall and rehearsal loop also enabled our ancestors to engage in a stream of thought. One thought or idea evokes another, revised version of it, which evokes yet another, and so forth recursively. In this way, attention is directed away from the external world toward one's internal model of it. Finally, self-triggered recall allowed for voluntary rehearsal and refinement of actions, enabling systematic evaluation and improvement of skills and motor acts.

## Computational Model

Donald's hypothesis is difficult to test directly, for if correct it would leave no detectable trace. It is, however, possible to computationally model how the onset of the capacity for recursive recall would affect the effectiveness, diversity, and open-endedness of ideas generated in an artificial society. This section summarizes how we tested Donald's hypothesis using an agent-based computational model of culture referred to as 'EVOlution of Culture', abbreviated EVOC. Details of the modeling platform are provided elsewhere (Gabora, 2008b, 2008c; Gabora & Leijnen, 2009; Leijnen & Gabora, 2009).

[1] The term *mimetic* is derived from "mime," which means "to act out."

**The EVOC World**. EVOC uses neural network based agents that (i) invent new ideas, (ii) imitate actions implemented by neighbors, (iii) evaluate ideas, and (iv) implement successful ideas as actions. Invention works by modifying a previously learned action using learned trends (such as that more overall movement tends to be good) to bias the invention process. The process of finding a neighbor to imitate works through a form of lazy (non-greedy) search. An imitating agent randomly scans its neighbors, and adopts the first action that is fitter than the action it is currently implementing. If it does not find a neighbor that is executing a fitter action than its own action, it continues to execute the current action. Over successive rounds of invention and imitation, agents' actions improve. EVOC thus models how descent with modification occurs in a purely cultural context. Agents do not evolve in a biological sense–they neither die nor have offspring–but do in a cultural sense, by generating and sharing ideas for actions.

Following Holland (1975) we refer to the success of an action in the artificial world as its *fitness,* with the caveat that unlike its usage in biology, here the term is unrelated to number of offspring (or ideas derived from a given idea). The fitness function (FF) was originally chosen because it allows investigation of biological phenomena such as underdominance and epistasis in a cultural context (see Gabora, 1995); the one used here is one over several used in EVOC (see Gabora, 2008 for others). The FF rewards head immobility and symmetrical limb movement. Fitness of actions starts out low because initially all agents are entirely immobile. Soon some agent invents an action that has a higher fitness than doing nothing, and this action gets imitated, so fitness in- creases. Fitness increases further as other ideas get invented, assessed, implemented as actions, and spread through imitation. The diversity of actions initially increases due to the proliferation of new ideas, and then decreases as agents hone in on the fittest actions.

We used was a toroidal lattice with 100 nodes, each occupied by a single, stationary agent, and a von Neumann neighborhood structure (agents only interacted with their four adjacent neighbors). During invention, the probability of changing the position of any body part involved in an action was 1/6. On each run, creators and imitators were randomly dispersed.

**Chaining.** This gives agents the opportunity to execute multi-step actions. For the experiments reported here with chaining turned on, if in the first step of an action an agent was moving at least one of its arms, it executes a second step, which again involves up to six body parts. If, in the first step, the agent moved one arm in one direction, and in the second step it moved the same arm in the opposite direction, it has the opportunity to execute a three-step action. And so on. The agent is allowed to execute an arbitrarily long action so long as it continues to move the same arm in the opposite direction to the direction it moved previously. Once it does not do so, the chained action comes to an end. The longer it moves, the higher the fitness of this multi-step chained action. Where $n$ is the number of chained actions, the fitness, $F_c$, is calculated as follows:

$$F_c = F_{nc} + (n - 1)$$

The fitness function with chaining provides a simple means of simulating the capacity for recursive recall.

## 'Origins of Creativity' Results

As shown in Figure 1, the capacity to chain simple actions into more complex ones increases the mean fitness of actions in the society. This is most evident in the later phase of a run. Without chaining, agents converge on optimal actions, and the mean fitness of action reaches a plateau. With chaining, however, there is no ceiling on the mean fitness of actions. By the 100$^{th}$ iteration it reached almost 15, indicating a high incidence of chaining.

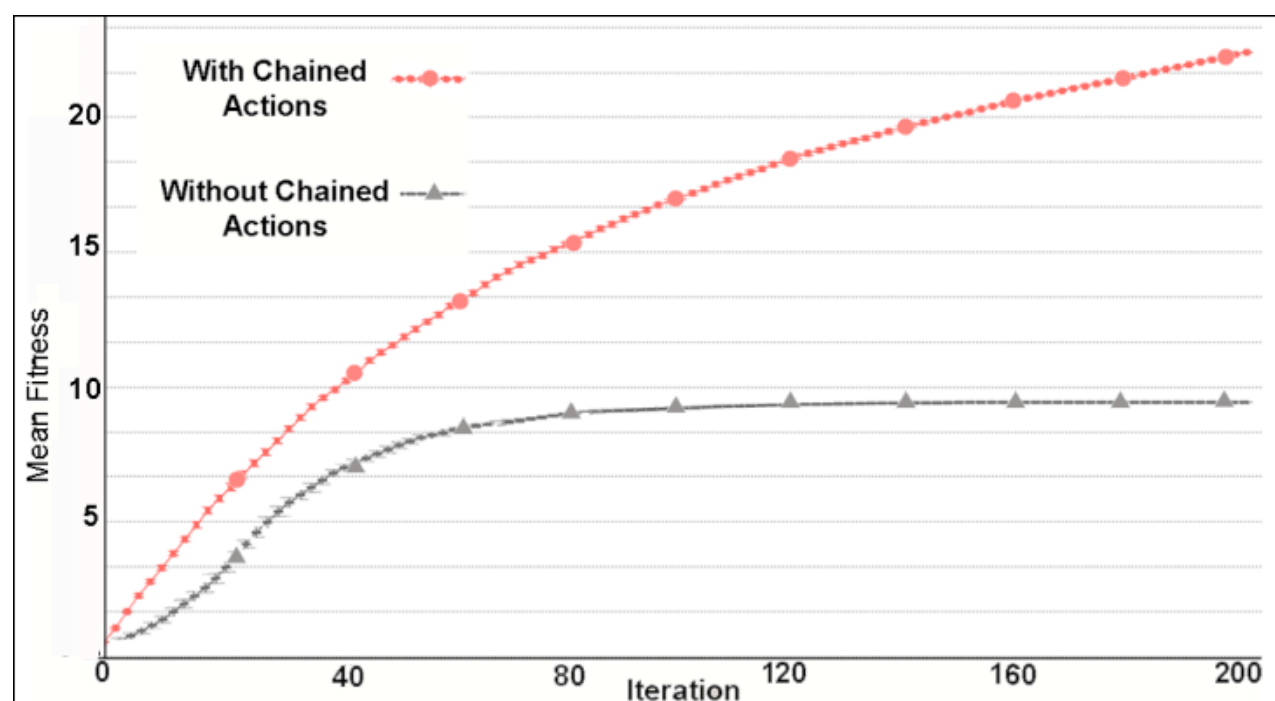

**Figure 1. Mean fitness of actions in the artificial society with chaining versus without chaining.**

As shown in Figure 2, chaining also increases the diversity of actions. This is most evident in the early phase of a run before agents begin to converge on optimal actions. Although in both cases there is convergence on optimal actions, without chained actions, this is a static set (thus mean fitness plateaus) whereas with chained actions the set of optimal actions is always changing, as increasingly fit actions are found (thus mean fitness keeps increasing).

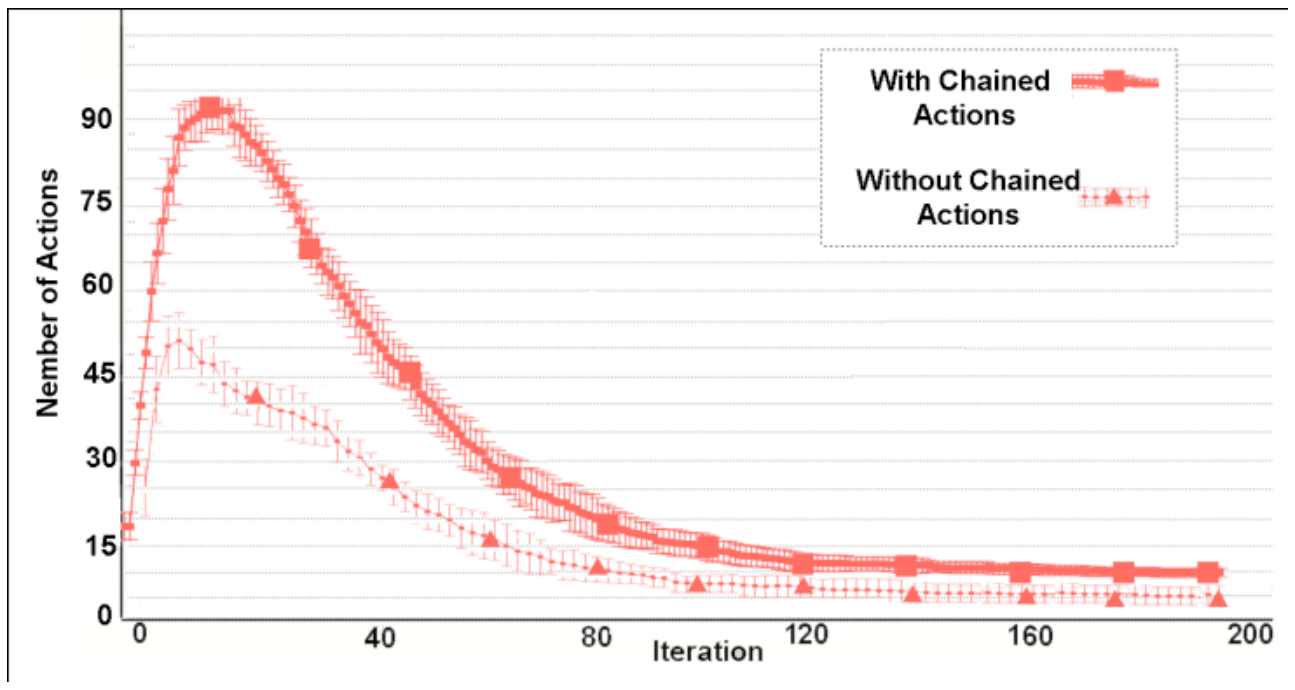

**Figure 2. Mean number of different actions in the artificial society with chaining (continuous line) versus without chaining (dashed line).**

Recall that agents can learn trends from past experiences (using the knowledge-based operators), and thereby bias the generation of novelty in directions that have a greater than chance probability of being fruitful. Since chaining provides more opportunities to capitalize on the capacity to learn, we hypothesized that chaining would accentuate the impact of learning on the mean fitness of actions, and this too turned out to be the case (Gabora & Saberi, 2011).

## Second Transition: The 'Big Bang' of Human Creativity

The European archaeological record indicates that a truly unparalleled cultural transition occurred between 60,000 and 30,000 years ago at the onset of the Upper Paleolithic (Bar-Yosef, 1994; Klein, 1989; Mellars, 1973, 1989a, 1989b; Soffer, 1994; Stringer & Gamble, 1993). Considering it "evidence of the modern human mind at work," Richard Leakey (1984:93-94) describes the Upper Palaeolithic as follows: "unlike previous eras, when stasis dominated, ... [with] change being measured in millennia rather than hundreds of millennia." Similarly, Mithen (1996) refers to the Upper Paleolithic as the 'big bang' of human culture, exhibiting more innovation than in the previous six million years of human evolution. We see the more or less simultaneous appearance of traits considered diagnostic of behavioral modernity. It marks the beginning of a more organized, strategic, season-specific style of hunting involving specific animals at specific sites, elaborate burial sites indicative of ritual and religion, evidence of dance, magic, and totemism, the colonization of Australia, and replacement of Levallois tool technology by blade cores in the Near East. In Europe, complex hearths and many forms of art appeared, including cave paintings of animals, decorated tools and pottery, bone and antler tools with engraved designs, ivory statues of animals and sea shells, and personal decoration such as beads, pendants, and perforated animal teeth, many of which may have indicated social status (White, 1989a, 1989b).

Whether this period was a genuine revolution culminating in behavioral modernity is hotly debated because claims to this effect are based on the European Palaeolithic record, and largely exclude the African record (McBrearty & Brooks, 2000); Henshilwood & Marean, 2003). Indeed, most of the artifacts associated with a rapid transition to behavioral modernity at 40–50,000 years ago in Europe are found in the African Middle Stone Age tens of thousands of years earlier. However the traditional and currently dominant view is that modern behavior appeared in Africa between 50,000 and 40,000 years ago due to biologically evolved cognitive advantages, and spread replacing existing species, including the Neanderthals in Europe (e.g., Ambrose, 1998; Gamble, 1994; Klein, 2003; Stringer & Gamble, 1993). Thus from this point onward there was only one hominid species: modern *Homo sapien*. Despite lack of overall increase in cranial capacity, the prefrontal cortex, and particularly the orbitofrontal region, increased significantly in size (Deacon, 1997; Dunbar, 1993; Jerison, 1973; Krasnegor, Lyon, and Goldman-Rakic, 1997; Rumbaugh, 1997) and it was likely a time of major neural reorganization (Klein, 1999; Henshilwood, d'Errico, Vanhaeren, van Niekerk, and Jacobs, 2000; Pinker, 2002).

Given that the Middle/Upper Palaeolithic was a period of unprecedented creativity, what kind of cognitive processes may have been involved?

## A Testable Hypothesis

Converging evidence suggests that creativity involves the capacity to shift between two forms of thought (Finke, Ward, & Smith, 1992; Gabora, 2003; Howard-Jones & Murray, 2003; Martindale, 1995; Smith, Ward, & Finke, 1995; Ward, Smith, & Finke, 1999). Divergent or associative processes are hypothesized to occur during idea generation, while convergent or analytic processes predominate during the refinement, implementation, and testing of an idea. It has been proposed that the Paleolithic transition reflects a mutation to the genes involved in the fine-tuning of the biochemical mechanisms underlying the capacity to subconsciously shift between these modes, depending on the situation, by varying the specificity of the activated cognitive receptive field. This is referred to as *contextual focus*[2] because it requires the ability to focus or defocus attention in response to the context or situation one is in. Defocused attention, by diffusely activating a broad region of memory, is conducive to divergent thought; it enables obscure (but potentially relevant) aspects of the situation thus come into play. Focused attention is conducive to convergent thought; memory activation is constrained enough to hone in and perform logical mental operations on the most clearly relevant aspects.

## Support from Computational Model

Again, because it would be difficult to empirically determine whether Paleolithic humans became capable of contextual focus, we decided to begin by determining whether the hypothesis is at least computational feasible. To do so we used an evolutionary art system that generated progressively evolving sequences of artistic portraits, with no human intervention once initiated. We sought to determine whether incorporating contextual focus into the fitness function would play a crucial role in enabling the computer system to generate art that humans find "creative" (i.e. possessing qualities of novelty and aesthetic value typically ascribed to the output of a creative artistic process).

We implemented contextual focus in the evolutionary art algorithm by giving the program the capacity to vary its level of fluidity and control over different phases of the creative process in response to the output it generated. The creative domain of portrait painting was chosen because it requires both focused attention and analytical thought to accomplish the primary goal of creating a resemblance to the portrait sitter, as well as defocused attention and associative thought to deviate from resemblance in a way that is uniquely interesting, *i.e.*, to meet the broad and often conflicting criteria of aesthetic art. Since judging creative art is subjective, rather than use quantitative analysis, a representative subset of the automatically produced artwork from this system was selected, output to high quality framed images, and submitted to peer reviewed and commissioned art shows, thereby allowing it to be judged positively or negatively as creative by human art curators, reviewers and the art gallery going public.

Our strategy for modeling contextual focus may raise questions about the ability of computers to "truly" be creative, and the role of the human system designer in the creative output. Several researchers in computational creativity, have addressed such questions by outlining different dimensions of creativity and proposing schema for evaluating a "level of creativity" of a given system, for example (Ritchie, 2007; Jennings, 2010; Colton, Pease, & Charnley, 2011). We are interested in applying such analyses to our portrait-system as a possibility for future work; indeed, the mechanics of contextual focus might be clarified by the computational creativity literature. In particular we are interested in further exploring the link between system-modified fitness constraints and the idea of transformational creativity (Boden, 2003; Wiggins 2006).

However, for the purposes of the current paper, it is less important to address the question of designer involvement in system creativity, or to try and quantify the amount of creativity displayed. Rather, we concentrate on the qualitative impact made by the explicit incorporation of contextual focus into the system as a whole, and its ability to elevate the perceived quality and novelty of system output to a level audiences judged reminiscent of successful "artistic, human-style" creativity.

**Generative Art Systems**: Creative evolutionary systems are a class of search algorithms inspired by Darwinian evolution, the most popular of which are genetic algorithms (GA) and genetic programming (GP) (Koza, 1993). These techniques solve complex problems by encoding a population of randomly generated potential solutions as 'genetic instruction sets', assessing the ability of each to solve the problem using a predefined fitness function, mutating and/or marrying (applying crossover to) the best to yield a new generation, and repeating until one of the offspring yields an acceptable solution. We are not claiming that contextual focus is Darwinian, but simply that for our computational modeling purposes, Genetic Programming proved a convenient foundational aggregator to support our contextual focus fitness function module.

Typically these systems allow a human user to pick those individuals that will be mated – making the human the creative judge. In contrast, our system used a function trigger mechanism within the contextual focus fitness func-tion which allowed the process to run automatically, without any human intervention once the process was started. It was not until the evolutionary art process came to completion that humans looked at and evaluated the art. Others have begun to use creative evolutionary systems with an automatic fitness function in design and music, as well as in a creative invention machine (Bentley, Corne, 2002). What is unique in our approach is that it incorporates several techniques that enable it to shift to processing artistic content in a more divergent or associative manner, and employs a form of GP called Cartesian Genetic Programming (Miller, 2011), detailed in the next section.

[2] In neural net terms, contextual focus amounts to the capacity to spontaneously vary the shape of the activation function, flat for divergent thought and spiky for analytical.

**Implementation:** The GP function set has 13 functions which use unitized *x* and *y* positions of the portrait image as variables and additional parameter variables (noted PM) that can be affected by adaptive mutation. Functions are low level in nature which aids in a large 'creative' search space, and output HSV color space values between 0 and 255. An individual in our population is manifested as one program that runs successively for every pixel in the output image, which is then tested against our creative fitness function. This allows correlated painterly effects as one moves through the image. Functions 1 through 5 use simple logical or arithmetic manipulations of the positions (low level functions create a larger 'creative' search space), whereas 7 through 14 use trigonometric or logical functions that are more related to geometric shapes and color graduations. The 13 functions of the function set are:

```
1: x|y;
2: PM & x;
3: (x ? y) % 255;
4: if (x[y) x - y; else y - x;
5: 255 - x;
6: abs (cos (x) * 255);
7: abs (tan (((x % 45) * pi)/180.0) * 255));
8: abs (tan (x) * 255) % 255);
9: sqrt ((x - PM)2 ? (y - PM) 2); (thresholded at 255)
10: x % (PM? 1) ? (255 - PM);
11: (x ? y)/2;
12: if (x[y)255*((y ? 1)/(x ? 1)); else 255*((x ? 1)/(y ? 1));
13: abs (sqrt (x – PM2? y – PM2) % 255);
```

The contextual focus based fitness function varies fluidly from tightly focusing on resemblance (similarity to the sitter image, which in this case is an image of Charles Darwin), to swinging (based on functional triggers) toward a more associative process of the intertwining, and at times contradicting, 'rules' of abstract portrait painting. Different genotypes map to the same phenotype. This allows us to vary the degree of creative fluidity because it offers the capacity to move though the search space via genotype (small ordered movement) or phenotype (large movement but still related). For example, in one set of experiments this is implemented as follows: if the fittest individual of a population is identical to an individual in the previous generation for more than three iterations, meaning the algorithm is stuck in analytic mode and needs to open up, other genotypes that map to this same phenotype are chosen over the current non-progressing genotype, allowing divergent open movement through the landscape of possibilities.

The automatic fitness function partly uses a 'portrait to sitter' resemblance. Since the advent of photography (and earlier), portrait painting has not just been about accurate reproduction, but also about using modern painterly goals to achieve a creative representation of the sitter. The fitness function primarily rewards accurate representation, but in certain situations also rewards visual painterly aesthetics using simple rules of art creation as well as a portrait knowledge space. Specifically, the divergent painterly portion of the fitness function takes into account: (1) face versus background composition, (2) tonal similarity over exact color similarity, matched with a sophisticated artistic color space model that weighs for warm-cool color temperature relationships based on analogous and complementary color harmony rules, and (3) unequal dominate and subdominant tone and color rules, and other artistic rules based on a portrait painter knowledge domain as detailed in (DiPaola, 2009) and illustrated in Figure 3. The system is biased toward resemblance, which gives it structure, but can, under the influence of functional triggers, exhibit artistic flair.

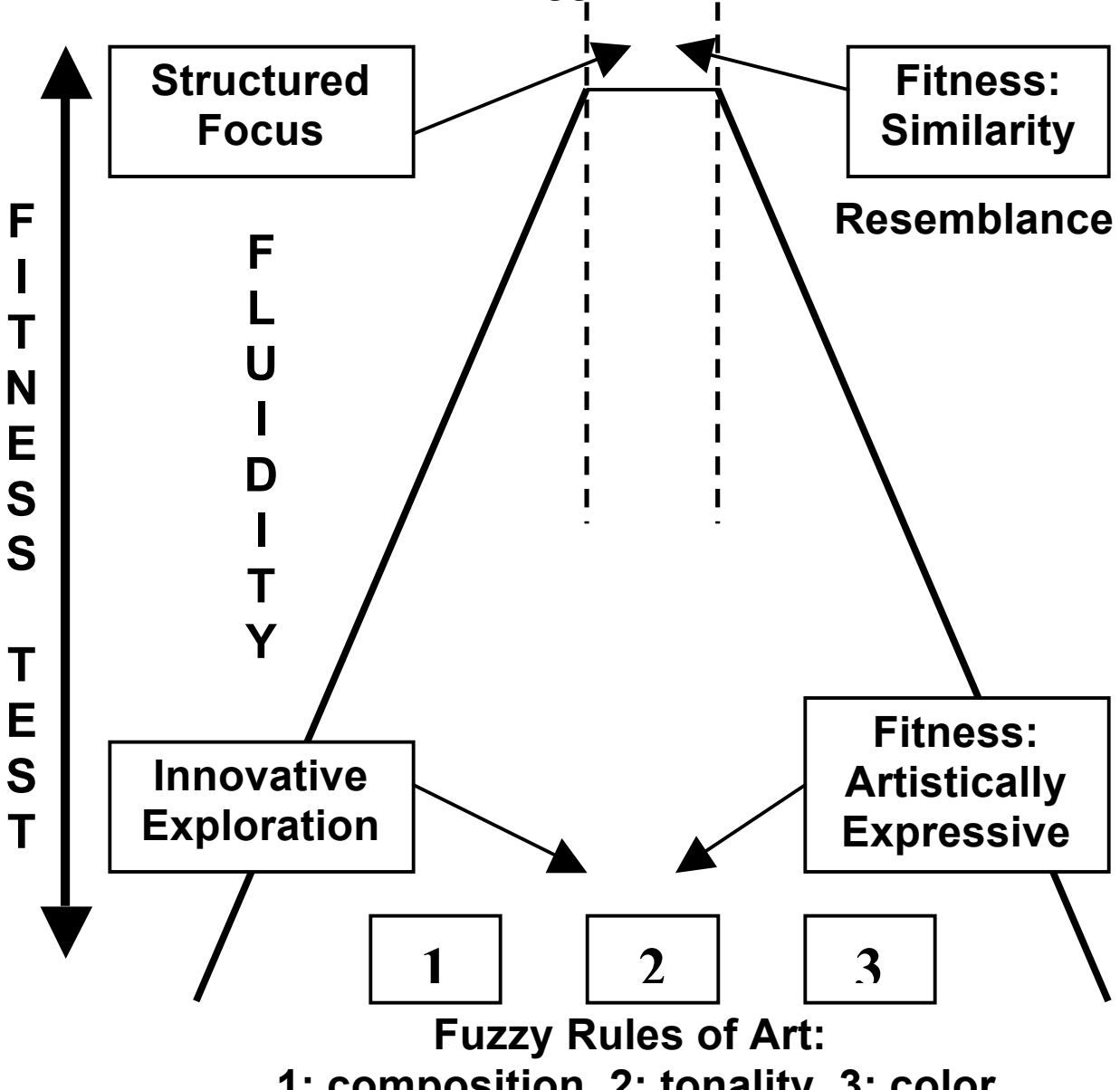

**Figure 3. The contextual focus fitness function mimics human creativity by moving between restrained focus (resemblance) to more unstructured associative focus (resemblance + ambiguous art rules of composition, tonality and color theory).**

The fitness function calculates four scores (resemblance and the three painterly rules), and then combines them in different ways to mimic human creativity, shifting between unstructured associative focus (rules of composition, tonality and color theory) and restrained focus (resemblance). In its default state, the fitness function uses a more analytic form of processing, specifically, a ratio of 80% resemblance to 20% non-proportional scoring of the three painterly rules. Several functional triggers can alter this ratio in different ways, but the main trigger is when the system is "stuck". Within any run, for instance as long as an adaptive percentage of 80–20 resemblance bias is maintained (resemblance patriarchs), the system will allow very high scoring of painterly rule individuals to be accepted into the next population. Those with high painterly scores (weighted non-proportionally including for a very high score with respect to just one rule) are saved separately, and mated with the current 80/20 population. Unless other triggers exist, their offspring are still tested with the 80–20 resemblance test. System wide functional changes occur when redundancy triggers affect the default ratio for all individuals. As mentioned previously, when a plateau or local minimum is reached for a certain number of populations, the fitness function ratio switches such that painterly rules are weighted higher than resemblance (on a sliding scale), and work in conjunction with redundancy at the input, node, and functional levels. Similarly, but now in reverse, to the default resemblance situation, high scoring resemblance individuals can pass into the next population when a percentage of painterly rule individuals is met. Using this more associative mode, high resemblance individuals are always part of the mix, and when these individuals show a marked improvement, a trigger is set to return to the more focused 80/20 resemblance ratio.

As the fitness score increases, portraits look more like the sitter. This gives us a somewhat known spread from very primitive (abstract) all the way through to realistic portraits. Thus in effect the system has two ongoing processes: (1) those 'most fit' portraits that pass on their portrait resemblance strategies, making for more and more realistic portraits—the family 'resemblance' patriarchs, and (2) the creative 'strange uncles': related to the current 'resemblance fit', but portraits that are more artistically creative or 'artistically fit'. This dual evolving technique of 'patriarchs and strange uncles' mimics the interplay between freedom and constraint that is so central to creativity. Paradoxically, novelty often benefits from the existence of a known framework reference system to rebel and innovate from. Creative people use some strong structural rules (as in the templates of a sonnet, tragedy, or in this case, a resemblance to the sitter image) as a resource or base to elaborate new variants beyond that structure (in this case, an abstracted variation of the sitter image).

## 'Big Bang of Creativity' Results

The automatic creative output was generated over thirty days of continuous, un-supervised computer use. The images in Figure 4 show a selection of representative portraits produced by the system. While the overall population improves at resembling Darwin's portrait, what is more interesting to us is the variety of recurring, emergent and merged creative strategies that evolve as the programs in different ways to become better abstract portraitists.

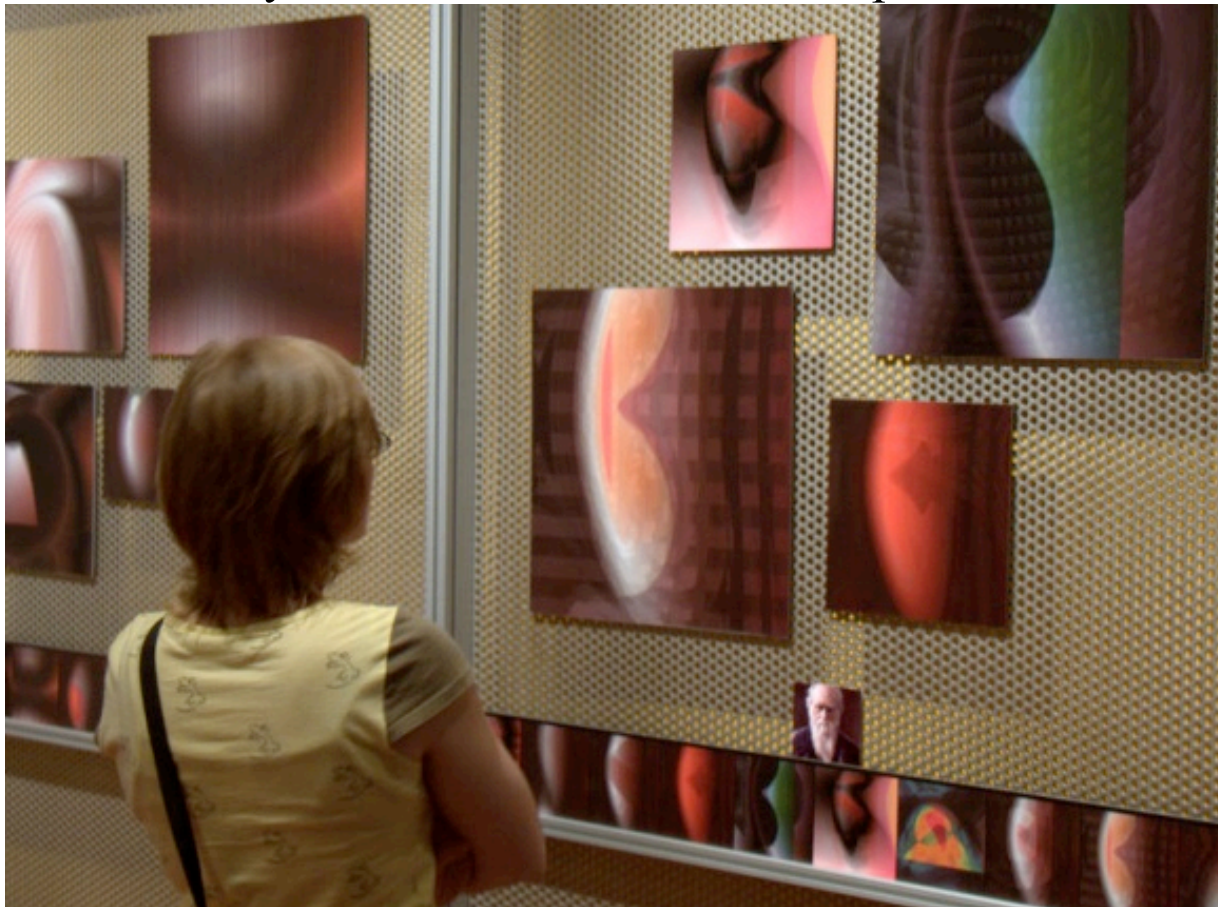

**Figure 4. These images have been seen by thousands in the last 2 years and have been perceived as creative art works on their own by the art public, including above at the MIT Museum in Cambridge, MA.**

Humans rated the portraits produced by this version of the portrait painting program with contextual focus as much more creative and interesting than a previous version that did not use contextual focus, and unlike its predecessor, the output of this program generated public attention worldwide. Example pieces were framed and submitted to galleries as a related set of work. Care was taken by the author to select representational images of the evolved unsupervised process, however creative human bias obvious exists in the representational editing process. Output has been accepted and exhibited at six major galleries and museums including the TenderPixel Gallery in London, Emily Carr Galley in Vancouver, and Kings Art Centre at Cambridge University as well as the MIT Museum, and the High Museum in Atlanta, all either peer reviewed, juried or commissioned shows from institutions that typically only accept human art work. A typical gallery installation consisted of 40-70 related portraits produced in time order over a given run. Gallery showings focus on "best resemblances" and those that are artistically compelling from an abstract portrait perspective. This gallery of work has been seen by tens of thousands of viewers who have commented that they see the artwork as an aesthetic piece that ebb and flows through seemly creative ideas even though they were solely created by an evolutionary art computer program using contextual focus. Note that no attempt to create a pure 'creativity Turning Test' was attempted. Besides the issues surrounding the validity of such a test (Pease, Colton, 2011), it was not feasible in such reputable and large art venues. However most of the thousands of causal viewers assumed they were looking at human created art. The work was also selected for its aesthetic value to accompany an opinion piece in the journal Nature (Padian, 2008), and was given a strong critical review by the Harvard humani-ties critic, Browne (2009). While these are subjective measures, they are standard in the art world. The fact that the computer program produced novel creative artifacts, both as single art pieces and as a gallery collection of pieces with interrelated themes, using contextual focus as a key element of its functioning, is compelling evidence of the effectiveness of contextual focus.

## Discussion and Conclusions

Many species engage in acts that could be said to be creative. However, humans are unique in that our creative ideas build on each other cumulatively; indeed it is for this reason that culture is widely construed as an evolutionary process (e.g. Bentley, Ormerod, & Batty, 2011; Cavalli-Sforza & Feldman, 1981; Gabora, 1996, 2008; Hartley, 2009; Mesoudi, Whiten & Laland, 2006; Whiten, Hinde, Laland, & Stringer, 2011). Our creativity is evident in all walks of life. It has transformed the planet we live on.

We discussed two transitions in the evolution of uniquely cumulative form of creativity, discussed cognitive mechanisms that have been proposed to underlie these transitions, and summarized efforts to computationally simulate them. Using an agent based computer model of cultural evolution, we obtained support for the hypothesis that the onset of cumulative, open-ended cultural evolution can be attributed to the evolution of a self-triggered recall and rehearsal loop, enabling the recursive chaining of thoughts and actions. Using a generative genetic programming system, we used a computational model of contextual focus to automatically produce a related series of art output that received critical acclaim usually given to human art work supporting the hypothesis that the capacity to shift between analytic and associative modes of thought plays an important role in the creative process.

Our results suggest that the evolution of chaining and contextual focus made possible the open-ended cumulative creativity exhibited by computational models of language evolution (e.g. Kirby, 2001). Note that in chaining versus no chaining conditions the size of the neural network is the same, but how it is used differs. This suggests that it was not larger brain size *per se* that initiated the onset of cumulative culture, but that larger brain size enabled episodes to be encoded in more detail, allowing more routes for reminding and recall, thereby facilitating recursive redescription of information encoded in memory (Karmiloff-Smith, 1992), thereby tailor it to the situation at hand. Our results suggest that it is reasonable to hypothesize that this in turn is vastly accentuated by the capacity to shift between associative and analytic different processing modes.

We wish to acknowledge some limitations of this work. Chaining does not work, as in humans, by considering an idea in light of one perspective, seeing how that perspective modifies the idea, seeing how this modification suggests a new perspective from which to consider the idea, and so forth. We are planning a more sophisticated implementation of that works more along these lines. Second, there is some irony in using an art program based on the genetic algorithm as a starting point to implement contextual focus, which we have claimed is unique to the cultural evolution of ideas and has no counterpart in biological evolution. Our goal here was to see if contextual focus 'works' at all; since this was successful, we will now move on to more cognitively plausible implementations. One of the projects currently underway is to implement contextual focus in the EVOC model of cultural evolution that was used for the 'origin of creativity' experiments. This is being carried out as follows. The fitness function will change periodically, so that agents find themselves no longer performing well. They will be able to detect that they are not performing well, and in response, increase the probability of change to any component of a given action. This temporarily makes them more likely to "jump out of a rut" resulting in a very different action, thereby simulating the capacity to shift to a more associative form of thinking. Once their performance starts to improve, the probability of change to any component of a given action will start to decrease to base level, making them less likely to shift to a dramatically different action. This helps them perfect the action they have already settled upon, thereby simulating the capacity to shift to a more associative form of thinking.

## Acknowledgements

We are grateful to Graeme McCaig and grants from *Natural Sciences and Engineering Research Council of Canada* and the Fund for Scientific Research of Flanders, Belgium.